\documentclass{article}

\usepackage{arxiv}

\usepackage[utf8]{inputenc} % allow utf-8 input
\usepackage[T1]{fontenc}    % use 8-bit T1 fonts
\usepackage{hyperref}       % hyperlinks
\usepackage{url}            % simple URL typesetting
\usepackage{booktabs}       % professional-quality tables
\usepackage{amsfonts}       % blackboard math symbols
\usepackage{nicefrac}       % compact symbols for 1/2, etc.
\usepackage{microtype}      % microtypography

\usepackage{graphicx} 
\usepackage{amsmath}
\usepackage{caption}
\usepackage{subcaption}

\title{Sequential Training Algorithm for Neural Networks}

\author{
  Jongrae Kim\thanks{http://robustlab.org; permanent email address: myjr52@gmail.com}\\
  Institute of Design, Robotics \& Optimisation (iDRO)\\
  Department of Mechanical Engineering\\
  University of Leeds\\
  Leeds LS2 9CT, UK\\
  \texttt{menjkim@leeds.ac.uk} \\
}

\DeclareMathOperator*{\minimise}{minimise}

\begin{document}
\maketitle

\begin{abstract}
A sequential training method for large-scale feedforward neural networks is presented.
Each layer of the neural network is decoupled and trained separately. After the training is completed
for each layer, they are combined together. The performance of the network would be sub-optimal compared
to the full network training if the optimal solution would be achieved. However, achieving
the optimal solution for the full network would be infeasible or require long
computing time. The proposed sequential approach
reduces the required computer resources significantly and would have better convergences as a single
layer is optimised for each optimisation step. The required modifications of existing algorithms to
implement the sequential training are minimal. The performance is verified by a simple example.
\end{abstract}

% keywords can be removed
\keywords{large-scale artificial neural network \and sequential training \and feedforward neural network}

\section{Introduction}
The artificial neural network has been applied to solve challenging problems in the past several years.
The feedforward neural network is one of the widely used networks in the artificial neural network.
As more computing power has been available, it is encouraged to apply the neural network to 
large-scale data sets.
The number of optimisation parameters increases as the dimension of training data set increases.
Optimisation solvers would require large computing power and resources to obtain reasonable solutions
for the neural networks. This is one of the challenges in the current applications of the neural network
algorithms.

The stochastic gradient method is frequently used for solving the optimisation problems.
Overview of optimisation methods for large scale machine learning algorithms are presented
in \cite{2016arXiv160604838B}, where mostly the stochastic gradient methods are discussed details in
terms of theoretical backgrounds and numerical behaviours. 
In \cite{8692438}, a feature selection algorithm is proposed so that the dimension of optimisation space is 
reduced. The methods in both are all training the whole network at the same time.

For each layer of the network, the input is transferred into a higher or the same dimensional space.
As long as the information in the input is conserved after the transformation, the training could
be done sequentially. This is the main motivation of the sequential training algorithm.
In the following, the algorithm is presented and the performance is demonstrated by a simple example.

\begin{figure}[ht]
        \centering
        \includegraphics[width=0.7\textwidth]{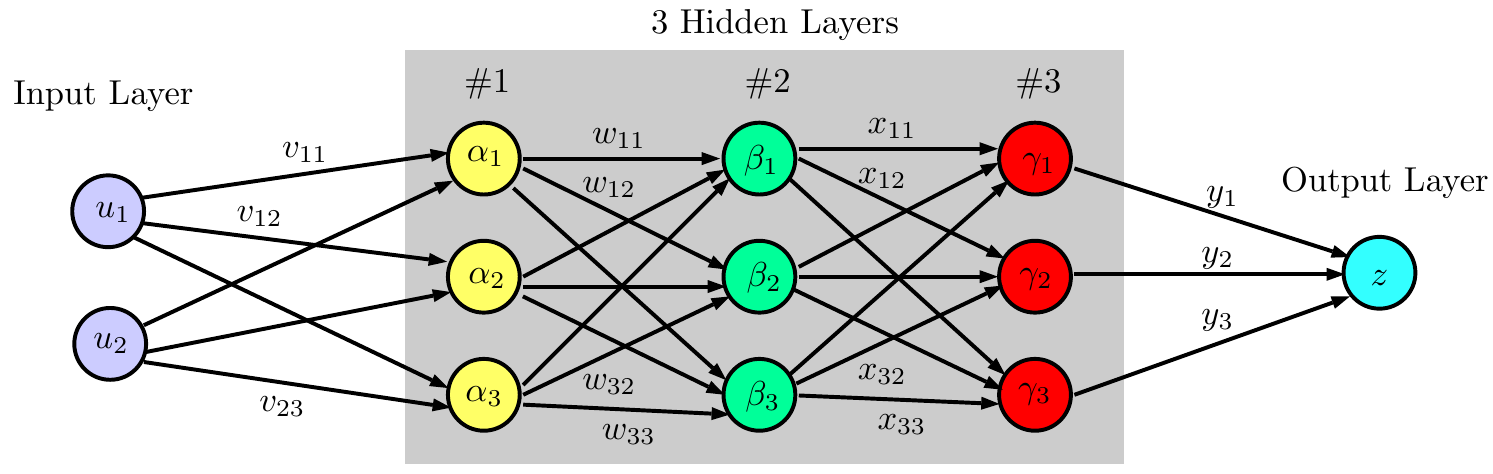}
        \caption{Full optimisation}
        \label{nn_3_layers}
\end{figure}

\section{Training Algorithm}

\subsection{Full Training}
Consider a neural network with 3 hidden layers shown in Figure \ref{nn_3_layers}.
The input and the output for the first layer are given by
\begin{align}
	{\boldsymbol \alpha} = {\bf f}(V {\bf u} + {\boldsymbol\theta})
\end{align}
where
\begin{align}
	{\boldsymbol \alpha} = \begin{bmatrix} \alpha_1\\ \alpha_2\\ \alpha_3 \end{bmatrix},~
	V = \begin{bmatrix} v_{11} & v_{12}\\ v_{21} & v_{22}\\ v_{31} & v_{32} \end{bmatrix},~
	{\bf u} = \begin{bmatrix} u_1\\ u_2 \end{bmatrix},~
	{\boldsymbol\theta} = \begin{bmatrix} \theta_1\\ \theta_2\\ \theta_3 \end{bmatrix},
\end{align}
$V$ and  ${\boldsymbol\theta}$ are the weight and the bias to be determined by an optimiser
in the first hidden layer, respectively, ${\bf f}(\cdot)$ is the activation function,
${\boldsymbol\alpha}$ is the output of the first hidden layer, and ${\bf u}$ is the input of the
training data set. 

Similarly, for the second hidden layer,
\begin{align}
	{\boldsymbol \beta} = {\bf f}(W {\boldsymbol\alpha} + {\boldsymbol\phi})
\end{align}
where 
\begin{align}
	{\boldsymbol \beta} = \begin{bmatrix} \beta_1\\ \beta_2\\ \beta_3 \end{bmatrix},~
	W = \begin{bmatrix} w_{11} & w_{12} & w_{13}\\ w_{21} & w_{22} & w_{23}\\ 
		w_{31} & w_{32} & w_{33} \end{bmatrix},~
	{\boldsymbol\phi} = \begin{bmatrix} \phi_1\\ \phi_2\\ \phi_3 \end{bmatrix},
\end{align}
${\boldsymbol\beta}$ is the output of the second hidden layer,
$W$ is  is the weight, and ${\boldsymbol\phi}$ is the bias for the second hidden layer. 
$W$ and ${\boldsymbol\phi}$ are design parameters to be obtained by an optimiser.

For the third hidden layer,
\begin{align}
	{\boldsymbol \gamma} = {\bf f}(X {\boldsymbol\beta} + {\boldsymbol\psi})
\end{align}
where 
\begin{align}
	{\boldsymbol \gamma} = \begin{bmatrix} \gamma_1\\ \gamma_2\\ \gamma_3 \end{bmatrix},~
	{\bf X} = \begin{bmatrix} x_{11} & x_{12} & x_{13}\\ x_{21} & x_{22} & x_{23}\\ 
		x_{31} & x_{32} & x_{33} \end{bmatrix},~
	{\boldsymbol\psi} = \begin{bmatrix} \psi_1\\ \psi_2\\ \psi_3 \end{bmatrix},
\end{align}
${\boldsymbol\gamma}$ is the output of the third hidden layer, and
$X$ and ${\boldsymbol\psi}$ are the weight and the bias to be determined, respectively.

Finally, the output of the network, $z$, is given by
\begin{align}
	z = f({\bf y}^T{\boldsymbol\gamma} + \omega)
\end{align}
where the superscript, $T$, is the transpose,
\begin{align}
	{\boldsymbol y}^T = \begin{bmatrix} y_1 &  y_2 & y_3 \end{bmatrix},~
\end{align}
${\bf y}$ and $\omega$ are the weight and the bias of the output node to be determined, respectively,
and $f(\cdot)$ is the activation function.\\

The following optimisation problem is solved to determine the unknowns:
\begin{align}\label{full_opt_prob}
	\minimise_{V, W, X, y, {\boldsymbol\theta}, {\boldsymbol\phi}, {\boldsymbol\psi}, \omega} 
	\sum_{i} \| z_i - z_i^* \|
\end{align}
where $z_i^*$ is the training data output. The total number of the unknowns in the minimisation problem
is 37(=6+9+9+3+3+3+3+1).

\begin{figure}[ht]
        \centering
        \includegraphics[width=0.95\textwidth]{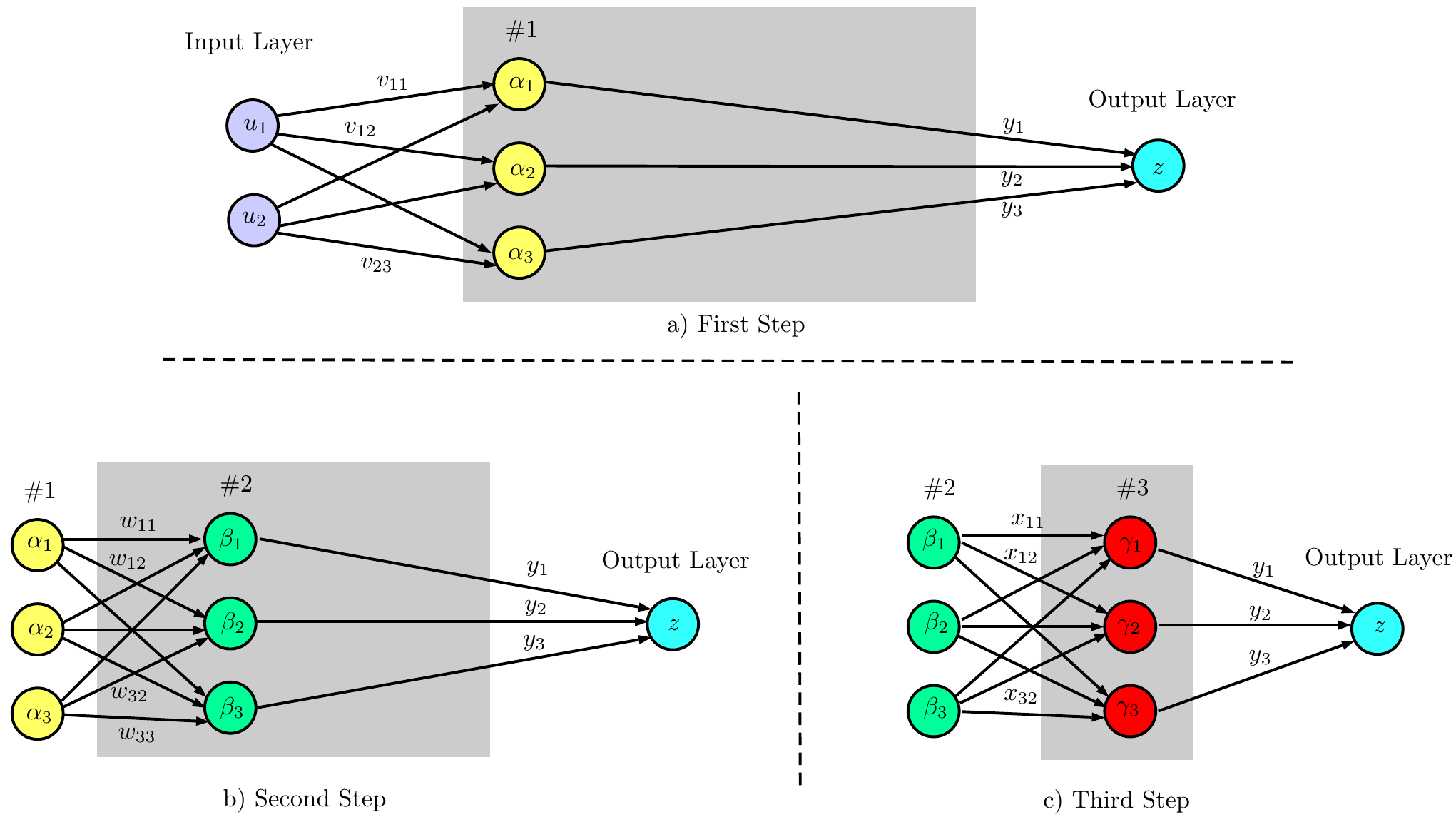}
        \caption{Sequential-optimisation}
        \label{nn_seq_opt}
\end{figure}

\subsection{Sequential Training}
\emph{Step 1:}
Instead of solving \eqref{full_opt_prob}, the full network is constructed sequentially as shown in 
Figure \ref{nn_seq_opt}. Firstly, introduce one hidden layer as shown in Figure \ref{nn_seq_opt}a),
where the following forward propagation is defined
\begin{subequations}
\begin{align}
	{\boldsymbol \alpha} &= {\bf f}(V u + {\boldsymbol\theta})\\
	z &= f({\bf y}^T{\boldsymbol\alpha} + \omega)
\end{align} 
\end{subequations}
and the optimisation problem to be solved is given by
\begin{align}\label{seq_opt_prob_step1}
	\minimise_{V, {\boldsymbol\theta}, {\bf y}, \omega} 
	\sum_{i} \| z_i - z_i^* \|
\end{align}
For the first step optimisation, the number of unknowns is 13(=6+3+3+1), which is almost one third of
the number of unknowns for the full optimisation. Once the first step optimisation is completed,
$V$ and ${\boldsymbol\theta}$ are determined. 
Two parameters in the output layer part in the first step optimisation,
i.e., ${\bf y}$ and $\omega$, are not used further but deserted.

\emph{Although ${\boldsymbol\alpha}$ is not enough to provide the perfect training 
	result from ${\bf u}$ to $z^*$, ${\boldsymbol\alpha}$ 
is still an improvement towards a better training. This can be, in turn, used as the training input together
with the training output, $z_i^*$, in the next step in order to obtain additional improvement in the training
result.}

\emph{Step 2:}
For the second forward propagation is defined by
\begin{subequations}
\begin{align}
	{\boldsymbol \beta} &= {\bf f}(W {\boldsymbol\alpha} + {\boldsymbol\phi})\\
	z &= f({\bf y}^T{\boldsymbol\beta} + \omega)
\end{align} 
\end{subequations}
as shown in Figure \ref{nn_seq_opt}b),
where
${\boldsymbol\alpha}$ obtained in the first step becomes the training input data for the second step 
optimisation. 

The optimisation problem to be solved is as follows:
\begin{align}\label{seq_opt_prob_step2}
	\minimise_{W, {\boldsymbol\phi}, {\bf y}, \omega} 
	\sum_{i} \| z_i - z_i^* \|
\end{align}
where the number of unknowns is 16(=9+3+3+1), which is less than a half of the number of unknowns 
for the full optimisation. 
Similar to the previous step, $W$ and ${\boldsymbol\phi}$ are determined,
${\boldsymbol\beta}$ becomes the training input data for the next step optimisation,
and ${\bf y}$ and $\omega$ are not used any further and deserted.

\begin{figure}[ht]
        \centering
	\begin{subfigure}[h]{0.479\textwidth}
	\centering
	\includegraphics[width=\textwidth]{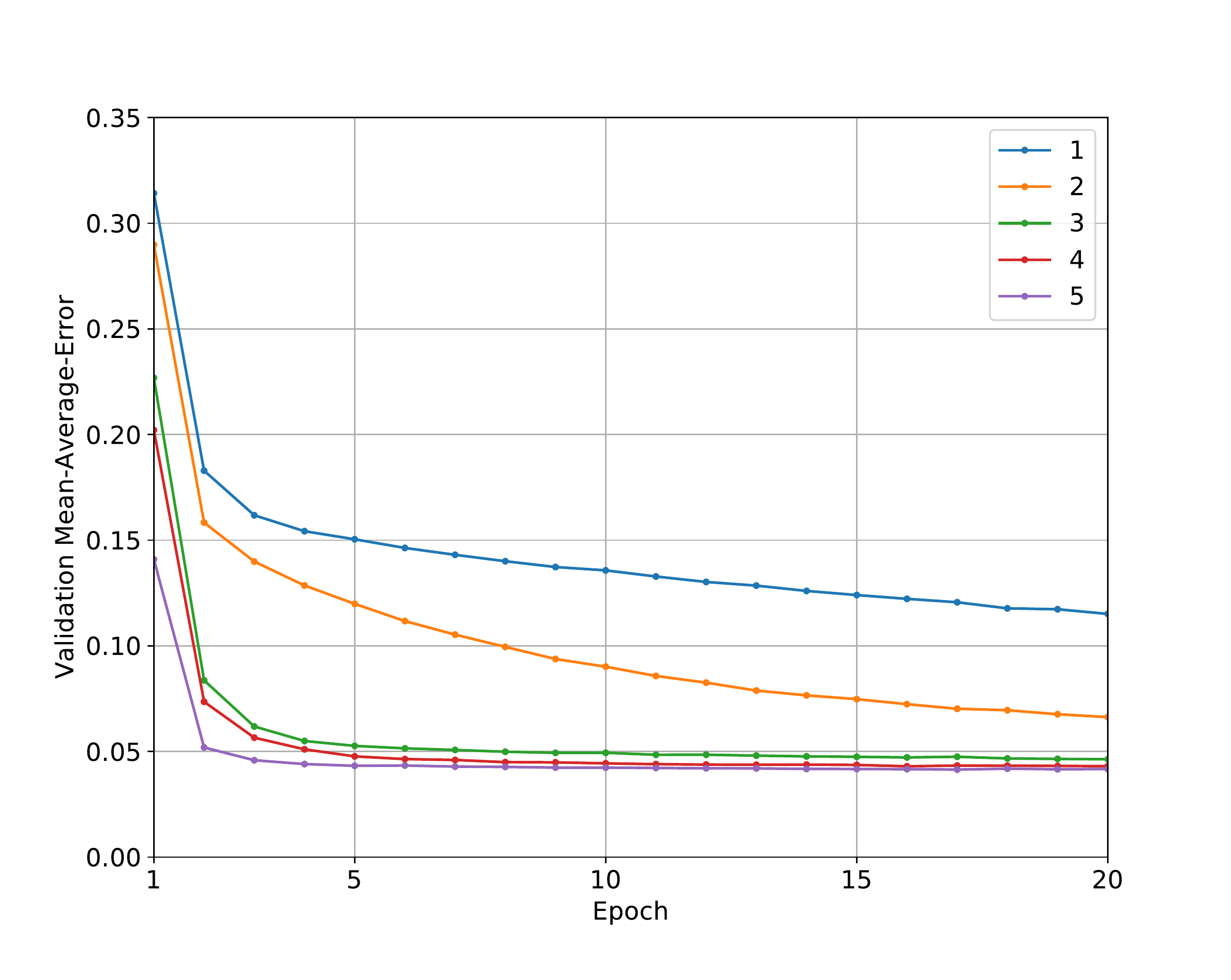}
	\caption{Mean error for each step in the sequential training}
        \label{seq_results}
	\end{subfigure}
	\begin{subfigure}[h]{0.495\textwidth}
	\centering
	\includegraphics[width=\textwidth]{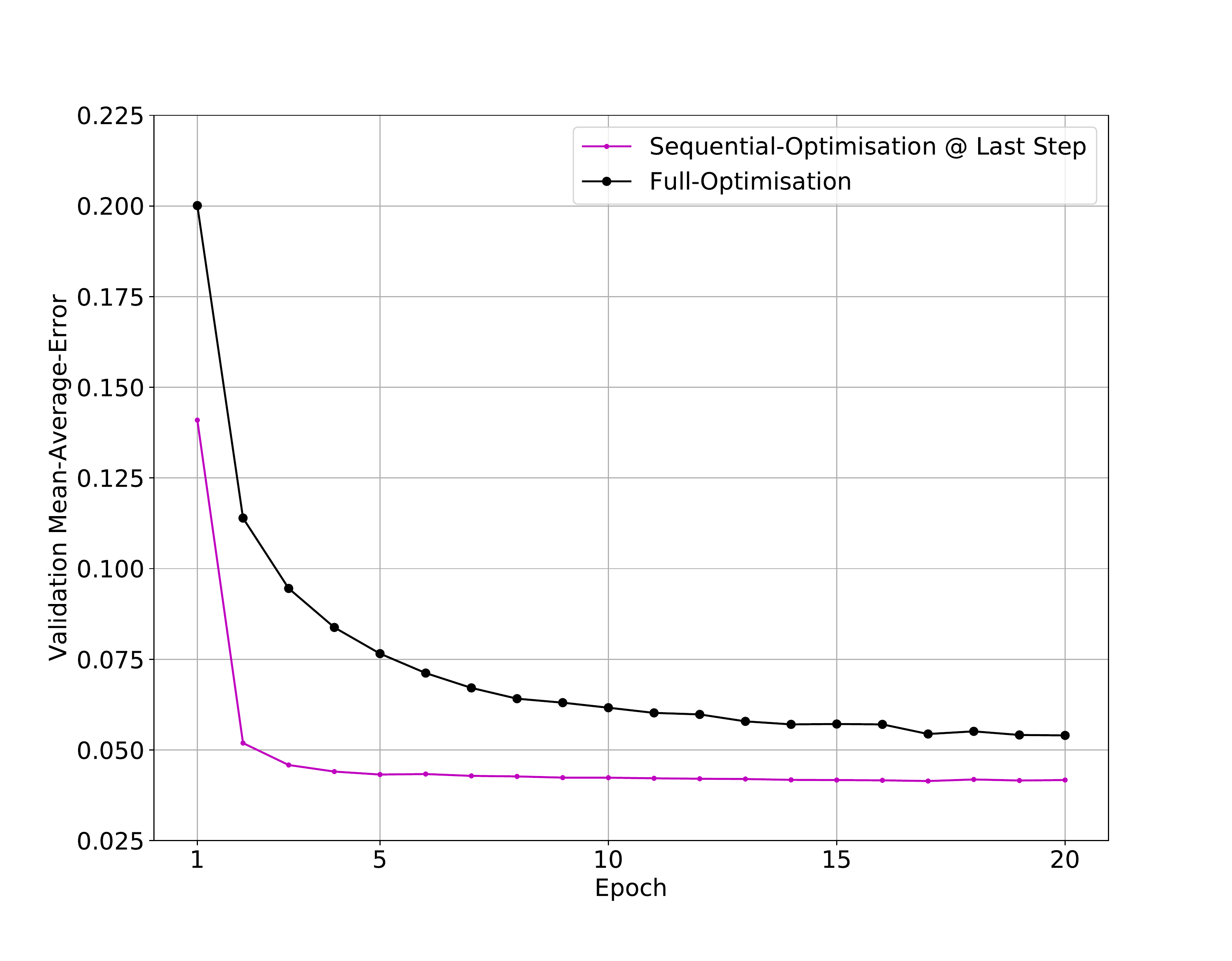}
	\caption{Mean errors for the full training and the sequential training}
        \label{seq_nn_compare_results}
	\end{subfigure}
	\caption{Sequential \& Full learning}
        \label{example_seq_full_results}
\end{figure}

\emph{Step 3:}
Finally, the third forward propagation is defined by
\begin{subequations}
\begin{align}
	{\boldsymbol \gamma} &= {\bf f}(X {\boldsymbol\beta} + {\boldsymbol\psi})\\
	z &= f({\bf y}^T{\boldsymbol\gamma} + \omega)
\end{align} 
\end{subequations}
as shown in Figure \ref{nn_seq_opt}c). The optimisation problem to be solved is as follows:
\begin{align}\label{seq_opt_prob_step3}
	\minimise_{X, {\boldsymbol\psi}, {\bf y}, \omega} 
	\sum_{i} \| z_i - z_i^* \|
\end{align}
where the number of unknowns is 16(=9+3+3+1), which is again less than a half of the number of unknowns 
for the full optimisation. Once the optimisation is completed, all parameters are determined including
${\bf y}$ and $\omega$ in the last step.

\emph{The reduction in the number of unknown parameters is significantly large as the number of hidden layers
and the dimension of training data increases.}

\section{Example}
The proposed sequential algorithm is tested with a simple example. A training data set is generated
using 
\begin{subequations}
\begin{align}
	{\bf u} &= \begin{bmatrix} x(0) & x(0)e^{2a} \end{bmatrix}^T,\\
	z^* &= a,
\end{align}
\end{subequations}
where $a$ is constant. 
The network to be trained provides an estimation of the system parameter, $a$, from
two measurements of $x(t)$ at $t=0$ and 2. Total 500 training data of $({\bf u}, a)$ are constructed by
500 random generations of $x(0)$ and $a$.

A neural network structure is set to have 5 hidden layers and each hidden layer has 16 nodes. 
The Keras is used to train the network with the ReLU(Rectified Linear Unit) 
activation function \cite{chollet2015keras}.
Figure \ref{seq_results} shows the average mean error of the validation for each step.
The error is gradually decreased as the step increases.
The validation error for the last step is compared with the ones for the full optimisation
in Figure \ref{seq_nn_compare_results}. The performances of both methods for test sets, 
which are not shown in here, are similar to each other.

In this example, the sequential training achieved a lower mean validation error 
than the full training.
This would imply that reducing the number of optimisation parameters would provide some advantages to
the optimisation solver to find a better solution. 

\section{Conclusions \& Future Work}
A sequential training algorithm for large-scale feedforward neural network is presented
and the performance is demonstrated using a simple example. The algorithm requires minimal changes
in existing training algorithms and could be deployed with little effort.
The algorithm needs to be tested for much larger data sets and the possibility
to apply the algorithm for various neural network structures such as CNN(Convolutional Neural Network),
RNN(Recurrent Neural Network) or GAN(Generative Adversarial Network) would be investigated in future research.

\bibliographystyle{unsrt}  
\bibliography{references} 

\end{document}